# Hope Speech Detection in Social Media English Corpora: Performance of Traditional and Transformer Models

**Luis Ramos**
Instituto Politécnico Nacional, Centro de Investigación en Computación, Mexico City, Mexico
lramos2020@cic.ipn.mx

**Hiram Calvo**
Instituto Politécnico Nacional, Centro de Investigación en Computación, Mexico City, Mexico
hcalvo@cic.ipn.mx

**Olga Kolesnikova**
Instituto Politécnico Nacional, Centro de Investigación en Computación, Mexico City, Mexico
kolesnikova@cic.ipn.mx

## Abstract

The identification of hope speech has become a promised NLP task, considering the need to detect motivational expressions of agency and goal-directed behaviour on social media platforms. This proposal evaluates traditional machine learning models and fine-tuned transformers for a previously split hope speech dataset as train, development and test set. On development test, a linear-kernel SVM and logistic regression both reached a macro-F1 of 0.78; SVM with RBF kernel reached 0.77, and Naïve Bayes hit 0.75. Transformer models delivered better results, the best model achieved weighted precision of 0.82, weighted recall of 0.80, weighted F1 of 0.79, macro F1 of 0.79, and 0.80 accuracy. These results suggest that while optimally configured traditional machine learning models remain agile, transformer architectures detect some subtle semantics of hope to achieve higher precision and recall in hope speech detection, suggesting that larges transformers and LLMs could perform better in small datasets.

## 1 Introduction

Hope speech detection is a task that has recently gained prominence in natural language processing (NLP) (Ahani et al., 2024b; Arif et al., 2024). The trend is fuelled by opportunities in social media, and community platforms where the identification of hopeful expression can guide timely actions to enhance user well-being(Balouchzahi et al., 2023).

Hope refers to the perceived capability to formulate methods to achieve set targets, and self-motivate through agentic thought to act on such methods(Snyder, 2002). Nonetheless, hope should not be confused with an optimistic attitude or simply a positive feeling, since it goes further than that as an emotion (Tash et al., 2024). Analysing hope clarify the goals and expectations of individuals,

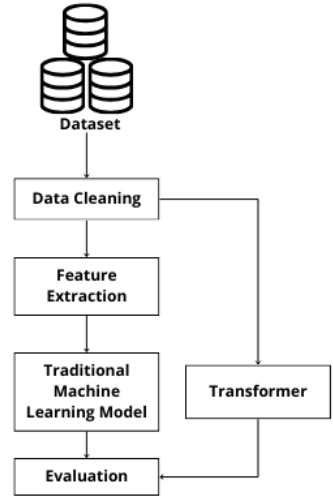

Figure 1: Overview of Methodology

societies, and even their subsets, such as genders and ethnicities(Balouchzahi et al., 2022).

In this paper, we propose a feature engineering framework for traditional machine learning (Ahani et al., 2024a; Ramos et al., 2025) and transformer models (Kolesnikova et al., 2025) for hope-speech detection, resulting in a comparison of classifiers performance. Systematic text cleaning standardizes misspellings, slang, and conceivable abbreviations employed commonly online (Nasiri and Budi, 2019). For traditional models, text representations functions were used such as TF-IDF and CountVectorizer from scikit-learn library with N-grams of words (Tash et al., 2025; Shahiki Tash et al., 2024).

On the development set, an SVM with linear kernel achieved a macro-F1 score of 0.78, closely followed by logistic regression at 0.78, SVM with RBF kernel at 0.77, and Naïve Bayes lagging slightly at 0.75. Concurrently, we fine-tuned two pre-trained transformer encoders: XLM-RoBERTa and RoBERTa-Dynabench. Both were trained for five epochs with batch size set to 16 under au-tomatic best-model selection based on macro-F1. On test-set results revealed that the RoBERTa-Dynabench model was on top with weighted precision 0.82, weighted recall 0.80, weighted F1 0.79, macro-precision 0.82, macro-recall 0.80, macro-F1 0.79 and overall accuracy 0.80, surpassing both the traditional SVM (0.79 across all metrics) and XLM-RoBERTa (0.79).

| Text | Label |
| --- | --- |
| #USER# Handsome king of Turkish God bless you and happiness in your life. InShallah God help you all the problems Ameenl | Hope |
| I just got gas for 4.89 and was almost happy about it. nnSorta hoping for another pandemic so I have to stay home. | Hope |
| I honest don't expect much in life, but I honestly dislike it when people treat my efforts as non-existence. :/ | Not Hope |
| lucas had every right to be skeptical like you bring a ranom ass girl from the woods and you expect me not to have some questions??? #URL# | Not Hope |

Table 1: Data Samples

These results reaffirm that traditional machine learning models like SVMs, despite performing well, are outperformed by transformer models which capture more deeply the complex semantic structures of hope expressions, enabling enhanced precision and recall in hope-speech recognition.

## 2 Literature Review

Hope speech detection is a novel area of research, and in this section we provide a review of different approaches in binary as well as multiclass classification datasets and techniques. This review aims to summarize the progress made so far and highlight the existing gaps in hope speech detection.

Recent years have seen an increasing interest in the computational analysis of hope speech. Chakravarthi, 2020 provided the first multilingual hope-speech corpus within the Equality, Diversity, and Inclusion framework, which includes 28 451 English, 20 198 Tamil, and 10 705 Malayalam YouTube comments. Comments were marked up with positive content, encouraging annotations representing different languages and cultures. The author experimented with various traditional machine learning models; however, all of them exhibited low performance.

García-Baena et al., 2023 created a dataset of 1,650 tweets divided into two categories: Hope Speech and Non-Hope Speech. Through their analysis, they showcased that a fine-tuned BETO surpassed both traditional (SVM, Naive Bayes, and Logistic Regression) and neural-based classifiers (MLP, CNN, and BiLSTM) when utilizing BERT-level embeddings with an F1-score of 85.12%.

Balouchzahi et al., 2023 recently proposed Poly-Hope, an English tweet corpus with a two-level classification of Hope, binary and a finer three-class subdivision into Generalized, Realistic, and Unrealistic Hope. Authors reported that transformers achieved a macro F1 score of 0.85 on the binary and 0.72 for the multiclass classification. In their subsequent work on Urdu language, Balouchzahi et al., 2025b created a five-class dataset that highlighted model divergence: Logistic Regression achieved the best for binary classification with a macro F1 score of 0.7593 and for multiclass classification, RoBERTa was the best with a macro F1 score of 0.4801.

Sidorov et al., 2024 brings forth two key contributions: building the first multiclass hope-speech corpus for Spanish and German based on tweets, and accomplishing a thorough assessment of detection frameworks across traditional machine learning, deep learning, and transformer models. Fine-tuned XLM-RoBERTa-base achieved an F1 score of 0.6801 on the Spanish data while UKLFR/GottBERT-base achieved 0.6977 on the German data, widely surpassing all non-transformer models.

Although all of these proposals contribute towards a coherent and important trend in the computational analysis of hope speech, they also suggest contexts wherein this analysis can be deepened and further developed.

## 3 Data Description

The dataset comprises English texts from social media. The dataset statistics are presented in Table 2, noting there is no imbalance between classes. Table 1 shows some samples from the dataset. The predictions were generated on the test set, which contains 2065 texts and was collected and split by the organizers of PolyHope-M shared task at RANLP 2025 (Balouchzahi et al., 2025a).

## 4 Methodology

In this section, we detail the proposed methodology to detect hope speech. In Figure 1 is presented an overview diagram for this proposal, and each phase is discrribed in detail in the following subsections.

| Classes | Train | Development |
|---|---|---|
| Hope | 2296 | 834 |
| Not Hope | 2245 | 816 |

Table 2: Data Statistics

### 4.1 Feature Engineering

This subsection expose the libraries and their functions used for feature engineering as a part of our machine learning pipelines. The tools facilitate data cleaning procedures as well as text vectorization techniques.

#### 4.1.1 Data Cleaning

This phase includes lower case, word lemmatization and deleting emojis, URLs, numbers, special characters, words between bracelets or number symbols like is shown in a Table 1 sample, and stop words. We used Spacy and NLTK libraries for lemmatization and deleting stop words respectively; Lemmatization helps reduce sparsity, increase generalization, and maximize the available information (Saraswati et al., 2025). This phase allows us to prepare the text for further analysis and modelling (Ridoy et al., 2024).

#### 4.1.2 Feature Extraction

In this phase we extract diverse features. Specially, we used two vectorization functions from scikit-learn library (Pedregosa et al., 2011; Ramos et al., 2024), *TfidfVectorizer* (TF-IDF) and *CountVectorizer* (CV). We tested diverse configuration parameters in these functions, such as unigrams, bigrams of words and characters, but we observed the best performance setting *ngram_range=(1,8)* and *analyzer='word'* for both functions in traditional machine learning models.

### 4.2 Traditional Machine Learning Models

Traditional machine learning models have shown consistent reference points, enabling reliable evaluation across various methods or algorithms (Sidorov et al., 2024). Experiments using traditional machine learning models have included diverse models such as Logistic Regression, Naïve Bayes and Support Vector Machine with linear and RBF kernel. Every model was deployed with default parameters using scikit-learn library (Pedregosa et al., 2011).

| Model | Vectorizer | Macro Precision | Macro Recall | Macro F1 |
|---|---|---|---|---|
| SVM-Linear | TF-IDF | 0.78 | 0.78 | 0.78 |
| SVM RBF | TF-IDF | 0.77 | 0.77 | 0.77 |
| Logistic Regression | CV | 0.79 | 0.78 | 0.78 |
| Naïve Bayes | TF-IDF | 0.75 | 0.75 | 0.75 |

Table 3: Best traditional machine learning model performances on development set

### 4.3 Transformer Models

Transformer architectures utilize self-attention mechanisms to dynamically capture context at different levels (Qasim et al., 2025). In hope speech detection have achieved promising results identifying hope expressions (Krasitskii et al., 2024). Experiments using transformers focused on BERT-based models. Specifically in two models from Hugging Face, *papluca/xlm-roberta-base-language-detection*[1] and *facebook/roberta-hate-speech-dynabench-r4-target*[2], which was trained for hate speech detection (Vidgen et al., 2020; Eponon et al., 2025). Both models were trained with the following specifications: 5 epochs, train and eval batch size of 16, *load_best_model_at_end = True* and *metric_for_best_model='f1'*. The last two parameters allow to choose the best model based on the best F1 score achieved for prediction.

## 5 Results

The macro F1 score is utilized as the key performance metric for comparison. On development set we only provide macro metrics but for the best model performances on test set we also provide macro precision and macro recall along with weighted precision, weighted recall, weighted F1 score and accuracy. Together, such metrics offer an all-rounded evaluation of the performance of the model.

### 5.1 Traditional Machine Learning Models

The Table 3 shows the comparison of the models performance on development set, as is showed, the best performance was achieved with SVM model using linear kernel and LR superpassing SVM with RBF kernel and NB, and we selected these models for prediction on test set.

---

[1] https://huggingface.co/papluca/xlm-roberta-base-language-detection
[2] https://huggingface.co/facebook/roberta-hate-speech-dynabench-r4-target

| Model | Macro Precision | Macro Recall | Macro F1 |
|---|---|---|---|
| distilbert/distilbert-base-uncased-finetuned-sst-2-english | 0.81 | 0.78 | 0.80 |
| papluca/xlm-roberta-base-language-detection | 0.80 | 0.77 | 0.79 |
| facebook/roberta-hate-speech-dynabench-r4-target | 0.79 | 0.81 | 0.80 |

Table 4: Best transformer performances on development set

### 5.2 Transformers Models

The Table 4 shows the comparison of the models performance on development set, as is showed, the best performance was achieved with *facebook/roberta-hate-speech-dynabench-r4-target* and *papluca/xlm-roberta-base-language-detection*. We selected these models for prediction on test set.

## 6 Discussion

Scores from the analysis reveal that the linear SVM and the *papluca/xlm-roberta-base-language-detection* framework settle around the same neighborhood-roughly 0.79-for precision, recall, and F1, each calculated on both a macro and a weighted basis. That outcome points to steady reliability across the classes the models encounter. In a separate vein, the *facebook/roberta-hate-speech-dynabench-r4-target* model nudges the numbers upward, hitting 0.82 for weighted precision and 0.80 for weighted recall; however, its weighted F1 plateaus at 0.79, so any improvement in coverage or sharpness stops short of generating a noticeably broader equilibrium between those counts.

## 7 Conclusion

This study had contributed to the research of hope speech detection through some key findings using traditional machine learning models and transformers. Results shows that machine learning models performance was close to the selected transformers models on test set. In summary, although traditional machine learning methods achieved comparable results that transformer-based systems, the transformers still surpassed their predecessors in precision and recall. This performance difference suggest that transformers outperforming traditional machine learning models detecting complex semantic patterns in hope speech, and also larges transformers and LLMs could perform better in small datasets.

| Model | Weighted Precision | Weighted Recall | Weighted F1 | Macro Precision | Macro Recall | Macro F1 | Acc |
|---|---|---|---|---|---|---|---|
| SVM-Linear | 0.79 | 0.78 | 0.79 | 0.79 | 0.79 | 0.79 | 0.79 |
| papluca/xlm-roberta-base-language-detection | 0.79 | 0.79 | 0.79 | 0.79 | 0.79 | 0.79 | 0.79 |
| facebook/roberta-hate-speech-dynabench-r4-target | 0.82 | 0.80 | 0.79 | 0.82 | 0.80 | 0.79 | 0.80 |

Table 5: Best model performances on test set